\documentclass[pdflatex,sn-mathphys]{sn-jnl}% Math and Physical Sciences Reference Style
\jyear{2021}%

\theoremstyle{thmstyleone}%
\theoremstyle{thmstyletwo}%

\theoremstyle{thmstylethree}%

\begin{document}

\title[Article Title]{Affect-driven Ordinal Engagement Measurement from Video}


\author*[1]{\fnm{Ali} \sur{Abedi}}\email{ali.abedi@uhn.ca}

\author[1]{\fnm{Shehroz} \sur{S. Khan}}\email{shehroz.khan@uhn.ca}

\affil[1]{\orgdiv{KITE}, \orgname{University Health Network, Canada}}


\abstract{In education and intervention programs, user engagement has been identified as a major factor in successful program completion. Automatic measurement of user engagement provides helpful information for instructors to meet program objectives and individualize program delivery. In this paper, we present a novel approach for video-based engagement measurement in virtual learning programs. We propose to use affect states, continuous values of valence and arousal extracted from consecutive video frames, along with a new latent affective feature vector and behavioral features for engagement measurement. Deep-learning sequential models are trained and validated on the extracted frame-level features. In addition, due to the fact that engagement is an ordinal variable, we develop the ordinal versions of the above models in order to address the problem of engagement measurement as an ordinal classification problem. We evaluated the performance of the proposed method on the only two publicly available video engagement measurement datasets, DAiSEE and EmotiW-EW, containing videos of students in online learning programs. Our experiments show a state-of-the-art engagement level classification accuracy of 67.4\% on the DAiSEE dataset, and a regression mean squared error of 0.0508 on the EmotiW-EW dataset. Our ablation study shows the effectiveness of incorporating affect states and ordinality of engagement in engagement measurement.}

\keywords{Engagement Measurement, Engagement Detection, Affect States, Temporal Convolutional Network, Ordinal Classification}



\maketitle

\section{Introduction}\label{sec1}
\label{sec:introduction}
Online services, such as virtual education, telemedicine, and telerehabilitation, offer many advantages compared to their traditional in-person counterparts; being more accessible, economical, and personalizable. These online services make it possible for students to complete their courses \cite{mukhtar2020advantages} and patients to receive the necessary care and health support \cite{nuara2021telerehabilitation} remotely. However, they also bring other types of challenges. For instance, in an online classroom setting, it becomes very difficult for the tutor to assess students' engagement in the material being taught \cite{venton2021strategies}. For a clinician, it is important to assess the engagement of patients in virtual intervention programs, as the lack of engagement is one of the most influential barriers to program completion \cite{matamala2020role}. Therefore, from the view of the instructor of an online program, it is important to automatically measure the engagement level of the participants to provide them with real-time feedback and take necessary actions to engage them to maximize their program's objectives.

Various modalities have been used for automatic engagement measurement in virtual settings, including user's image \cite{delgado2021student}, video \cite{abedi2021improving}, audio \cite{fedotov2018multimodal}, Electrocardiogram (ECG) \cite{belle2012automated}, and pressure sensors \cite{rivas2021multi}. Video cameras/webcams are mostly used in virtual programs; thus, they have been extensively used in assessing engagement in these programs. Video cameras/webcams offer a cheaper, ubiquitous and unobtrusive alternative to other sensing modalities. Therefore, majority of the recent works on objective engagement measurement in online programs and human-computer interaction are based on the visual data of participants acquired by cameras and using computer-vision techniques \cite{doherty2018engagement,dewan2019engagement,salam2022automatic,khan2022inconsistencies}.

The computer-vision-based approaches for engagement measurement are categorized into image-based and video-based approaches. The former approach measures engagement based on single images \cite{delgado2021student} or single frames extracted from videos \cite{whitehill2014faces}. A major limitation of this approach is that it only utilizes spatial information from single frames, whereas engagement is a spatio-temporal state that takes place over time \cite{d2017advanced}. Another challenge with the frame-based approach is that annotation is needed for each frame \cite{ringeval2013introducing}, which is an arduous task in practice. The latter approach is to measure engagement from videos instead of using single frames. In this case, one label is needed after each video segment. Fewer annotations are required in this case; however, the measurement problem is more challenging due to the coarse labeling.

The video-based engagement measurement approaches can be categorized into end-to-end and feature-based approaches. In the end-to-end approaches, consecutive raw frames of video are fed to variants of Convolutional Neural Networks (CNNs) (many times followed by temporal neural networks) to output the level of engagement \cite{abedi2021improving,liao2021deep,hu2022optimized,gupta2016daisee,zhang2019novel}. In feature-based approaches, handcrafted features, as indicators of engagement, are extracted from video frames and analyzed by sequential neural networks or machine-learning methods to output the engagement level \cite{copur2022engagement,whitehill2014faces,liao2021deep,niu2018automatic,thomas2018predicting,huang2019fine,fedotov2018multimodal,chen2019faceengage,booth2017toward,kaur2018prediction,wu2020advanced,ma2021automatic}. Various features have been used in previous feature-based approaches, including behavioral features such as eye gaze, head pose, and body pose. Despite the psychological evidence for the effectiveness of affect states in user engagement \cite{d2017advanced,sinatra2015challenges,khan2022inconsistencies,altuwairqi2021new}, none of the previous computer-vision-based approaches have utilized these important features for engagement measurement.

In this paper, we propose a novel feature-based approach for engagement measurement from videos using affect states, along with a new latent affective feature vector and behavioral features. These features are analyzed using different ordinal sequential neural networks to output user engagement in videos. Our main contributions are as follows:
\begin{itemize}
    \item We use affect states, including continuous values of user's valence and arousal, and present a new latent affective feature using a pretrained model \cite{toisoul2021estimation}, along with behavioral features for video-based engagement measurement. We jointly train different models on the above features for engagement measurement.
    \item In view of the fact that engagement is an ordinal variable \citep{whitehill2014faces,khan2022inconsistencies}, we approach engagement measurement from an ordinal classification perspective and train and evaluate ordinal classification models for engagement measurement.
    \item We conduct extensive experiments on two publicly available video datasets for engagement level classification and regression, focusing on studying the impact of affect states and ordinality of the classification models on engagement measurement. In these experiments, we compare the proposed method with several end-to-end and feature-based engagement measurement approaches.
\end{itemize}
The paper is structured as follows. Section \ref{sec:what_is_engagement} describes what engagement is and which type of engagement will be explored in this paper. Section \ref{sec:literature_review} describes related works on video-based engagement measurement. In Section \ref{sec:method}, the proposed pathway for affect-driven video-based ordinal engagement measurement is presented. Section \ref{sec:experimental_results} describes the experimental settings and results on the proposed methodology. In the end, Section \ref{sec:conclusion} presents our conclusions and directions for future works.

\section{What is Engagement?}
\label{sec:what_is_engagement}
In this section, different theoretical, methodological categorizations, and conceptualizations of engagement are briefly described, followed by brief discussion on the engagement in virtual learning settings and human-computer interaction which will be explored in this paper.

Dobrian et al. \cite{dobrian2011understanding} describe engagement as a proxy for involvement and interaction. Sinatra et al. \cite{sinatra2015challenges} defined "grain size" for engagement, the level at which engagement is conceptualized, observed, and measured. The grain size ranges from macro-level to micro-level. The macro-level engagement is related to groups of people, such as the engagement of employees of an organization in a particular project, the engagement of a group of older adult patients in a rehabilitation program, and the engagement of students in a classroom, school, or community. The micro-level engagement links with individual's engagement at the moment, task, or activity \cite{sinatra2015challenges}. Measurement at the micro-level may use physiological and psychological indices of engagement such as blink rate \cite{ranti2020blink}, head pose \cite{chen2019faceengage}, and heart rate \cite{monkaresi2016automated}.

The grain size of engagement analysis can also be considered as a continuum comprising context-oriented, person-in-context, and person-oriented engagement \cite{sinatra2015challenges}. The macro-level and micro-level engagement are equivalent to the context-oriented, and person-oriented engagement, respectively, and the person-in-context lies in between. The person-in-context engagement can be measured by observations of a person’s interactions with a particular contextual environment, e.g., reading a web page or participating in an online classroom. Except for the person-oriented engagement, for measuring the two other parts of the continuum, knowledge about context is required, e.g., the lesson being taught to the students in a virtual classroom \cite{sumer2021multimodal}, or the information about rehabilitation consulting material being given to a patient and the severity of disease of the patient in a virtual rehabilitation session \cite{rivas2021multi}.

In person-oriented engagement, the focus is on the behavioral, affective, and cognitive states of the person in the moment of interaction \cite{d2017advanced}.
\begin{itemize}
    \item Behavioral engagement involves general on-task behavior and paid attention at the surface level \cite{sinatra2015challenges, d2017advanced, aslan2017human}. The indicators of behavioral engagement, in the moment of interaction, include eye contact, blink rate, body pose, and hand pose.
    \item Affective engagement is defined as the affective and emotional reactions of the person to the content \cite{pekrun2012academic}. Its indicators are activating versus deactivating and positive versus negative emotions \cite{broughton2010nature}. Activating emotions are associated with engagement \cite{sinatra2015challenges}. While both positive and negative emotions can facilitate engagement and attention, research has shown an advantage for positive emotions over negative ones in upholding engagement \cite{broughton2010nature,altuwairqi2021new}.
    \item Cognitive engagement pertains to the psychological investment and effort allocation of the person to deeply understand the context materials \cite{sinatra2015challenges}. To measure cognitive engagement, information such as person’s speech should be processed to recognize the level of person’s comprehension of the context. Contrary to behavioral and affective engagements, measuring cognitive engagement requires knowledge about context materials.
\end{itemize}

Understanding a person’s engagement in a specific context depends on the knowledge about the person and the context. From a data analysis perspective, it depends on the data modalities available to analyze. In this paper, the focus is on video-based engagement measurement. The only data modality is video, without audio, and with no knowledge about the context. Therefore, we propose new methods for automatic person-oriented engagement measurement by analyzing the behavioral and affective states of the person at the moment of participation in an online program.

\section{Literature Review}
\label{sec:literature_review}
Over the recent years, extensive research efforts have been devoted to automatic engagement measurement \cite{doherty2018engagement,dewan2019engagement,salam2022automatic,khan2022inconsistencies}, most of which focused on video-based engagement measurement. In these approaches, computer vision, machine learning, and deep-learning algorithms are used to analyze videos and output the engagement level of the person in the video. Engagement measurement approaches based on video can be classified into end-to-end and feature-based.

In end-to-end engagement measurement approaches, no hand-crafted features are extracted from videos. The consecutive raw frames of RGB videos are fed to a deep CNN (or its variant) classification, or regression model to output a class label, or a continuous value corresponding to the engagement level of the person in the video. The upper half of Table \ref{tab:literature_review} summarizes the literature on the end-to-end video-based engagement measurement approaches focusing on their deep-learning video-analysis models. Most of the existing end-to-end approaches evaluated their method on the DAiSEE dataset \cite{gupta2016daisee} (described in Section \ref{sec:experimental_results}) for engagement level classification. Various techniques, such as using different loss functions or adding attention mechanisms to the neural networks, have been employed by researchers to enhance the results of their proposed methodologies \cite{abedi2021improving}.

Different from the end-to-end approaches, in feature-based approaches, first, multi-modal handcrafted features are extracted from videos, and then the features are fed to a classification or regression model to output engagement \cite{chen2019faceengage, doherty2018engagement, dewan2019engagement, whitehill2014faces, niu2018automatic, thomas2018predicting, huang2019fine, fedotov2018multimodal, dhall2020emotiw, booth2017toward, kaur2018prediction, wu2020advanced}. The lower half of Table \ref{tab:literature_review} summarizes the literature on feature-based engagement measurement approaches focusing on their features, feature fusion techniques, machine-learning models, and datasets.

As can be seen in the lower section of Table \ref{tab:literature_review}, in some of the previous methods, conventional computer-vision features such as box filter, Gabor \cite{whitehill2014faces}, or LBP-TOP \cite{kaur2018prediction} are used for engagement measurement. In some other methods, such as \cite{chen2019faceengage}, for extracting facial embedding features, a CNN containing convolutional layers, followed by fully-connected layers, is trained on a face recognition or facial expression recognition dataset. Then, the output of the convolutional layers in the pretrained model is used as the facial embedding features. In most of the previous methods, facial action units (AUs), eye movement, gaze direction, and head pose features are extracted using OpenFace \cite{baltrusaitis2018openface}, or body pose features are extracted using OpenPose \cite{cao2017realtime}. Various features are extracted and concatenated using these feature extraction techniques to construct a feature vector for each video frame or each video clip.

Late fusion and early fusion techniques were used in the previous feature-based engagement measurement approaches to handle different feature modalities. Late fusion requires the construction of independent models each being trained on one feature modality. For instance, Wu et al. \cite{wu2020advanced} trained four models independently, using face and gaze features, body pose features, and C3D features. Then they used a weighted summation of the outputs of four models to output the final engagement level. In the early fusion approach, features of different modalities are simply concatenated to create a single feature vector to be fed to a single model to output engagement. For instance, Niu et al. \cite{niu2018automatic}, concatenated gaze, AU, and head pose features to create one feature vector for each video clip and fed the feature vectors to a Gated Recurrent Unit (GRU) to output the engagement level.

While there is psychological evidence for the big influence of affect states on engagement \cite{sinatra2015challenges,d2017advanced, woolf2009affect, aslan2017human, altuwairqi2021new}, none of the previous methods utilized affect states as the features in engagement measurement. In the next section, we will present a new approach for video engagement measurement using affect states along with latent affective features, blink rate, eye gaze, head pose, and hand pose features. Despite the fact that engagement is an ordinal variable \citep{whitehill2014faces,khan2022inconsistencies}, none of the previous methods developed ordinal models for engagement measurement. For the first time in the field of affective computing, we use ordinal models for engagement level measurement.

\begin{table}
\centering
\caption{
(Upper section) End-to-end video-based engagement measurement approaches, their video-analysis classification (C) or regression (R) models, and the datasets they used. (Lower section) Feature-based video engagement measurement approaches, their features, feature fusion techniques, classification (C) or regression (R) models, and the datasets they used.
}
\resizebox{\columnwidth}{!}{
\begin{tabular}{p{.125\linewidth}p{.35\linewidth}p{.1\linewidth}p{.25\linewidth}p{.06\linewidth}p{.20\linewidth}}
 \hline
 Ref., Year & Features & Fusion & Model & C/R & Dataset \\
 \hline\hline
 \cite{gupta2016daisee}, 2016 & - & - & InceptionNet, C3D, and LRCN & C & DAiSEE \cite{gupta2016daisee}\\ 
 \hline
 \cite{geng2019learning}, 2019 & - & - & C3D & C & DAiSEE \cite{gupta2016daisee}\\ 
 \hline
 \cite{zhang2019novel}, 2019 & - & - & Inflated 3D-CNN (I3D) & C & DAiSEE \cite{gupta2016daisee}\\ 
 \hline
 \cite{liao2021deep}, 2021 & - & - & ResNet-50 with attention-based LSTM & C, R & DAiSEE \cite{gupta2016daisee}, EmotiW-EW \cite{kaur2018prediction}\\
 \hline
 \cite{abedi2021improving}, 2021 & - & - & ResNet with TCN & C & DAiSEE \cite{gupta2016daisee}\\
 \hline
 \cite{hu2022optimized}, 2022 & - & - &  ShuffleNet v2 & C & DAiSEE \cite{gupta2016daisee}\\
%  \hline
%  \cite{selim2022students}, 2022 & - & - &  EfficientNetB7 With TCN and LSTM & C & DAiSEE \cite{gupta2016daisee}\\
 \hline
 \cite{ai2022class}, 2022 & - & - &  Video Transformer & R & DAiSEE \cite{gupta2016daisee}, EmotiW-EW \cite{kaur2018prediction}\\
 \hline
 \cite{mehta2022three}, 2022 & - & - &  3D DenseNet & C, R & DAiSEE \cite{gupta2016daisee}, EmotiW-EW \cite{kaur2018prediction}\\

 \hline\hline
 
 \cite{whitehill2014faces}, 2014 & box filter, Gabor, AU & late & GentleBoost, SVM, logistic regression & C, R & HBCU \cite{whitehill2014faces}\\
 \hline
 \cite{booth2017toward}, 2017 & facial landmark, AU, optical flow, head pose & late & SVM, KNN, Random Forest & C, R & USC \cite{booth2017toward}\\
 \hline 
 \cite{kaur2018prediction}, 2018 & LBP-TOP & - & fully-connected neural network & R & EmotiW-EW \cite{kaur2018prediction}\\
 \hline 
 \cite{niu2018automatic}, 2018 & gaze direction, AU, head pose & early & GRU & R & EmotiW-EW \cite{kaur2018prediction}\\
 \hline
 \cite{fedotov2018multimodal}, 2018 & body pose, facial embedding, eye features, speech features & early & logistic regression & C & EMDC \cite{fedotov2018multimodal}\\
 \hline
 \cite{thomas2018predicting}, 2018 & gaze direction, head pose, AU & early & TCN & R & EmotiW-EW \cite{kaur2018prediction}\\
 \hline
 \cite{huang2019fine}, 2019 & gaze direction, head pose, AU & early & LSTM & C & DAiSEE \cite{gupta2016daisee}\\
 \hline
 \cite{chen2019faceengage}, 2019 & gaze direction, blink rate, head pose, facial embedding & early & AdaBoost, SVM, KNN, Random Forest, RNN & C & FaceEngage \cite{chen2019faceengage}\\
 \hline
 \cite{wu2020advanced}, 2020 & gaze direction, head pose, body pose, C3D & late & LSTM, GRU & R & EmotiW-EW \cite{kaur2018prediction}\\
 \hline
 \cite{ma2021automatic}, 2021 & gaze direction, head pose, AU, C3D & late & Neural Turing Machine & C & DAiSEE \cite{gupta2016daisee}\\
 \hline
 \cite{copur2022engagement}, 2022 & eye gaze, head pose, head
 rotation, and AU & early &  LSTM & R & EmotiW-EW
 \cite{kaur2018prediction}\\
 \hline
 \textbf{proposed} & \textbf{valence, arousal, latent affective features, blink rate, gaze direction, head pose, and hand pose} & \textbf{joint} & \textbf{Ordinal LSTM and TCN} & \textbf{C, R} & \textbf{DAiSEE \cite{gupta2016daisee}, EmotiW-EW \cite{kaur2018prediction}}\\
 \hline
\end{tabular}
}
\label{tab:literature_review}
\end{table}

\section{Affect-driven Ordinal Engagement Measurement from Video}
\label{sec:method}
As discussed in Section \ref{sec:what_is_engagement}, person-oriented engagement focuses on the behavioral, affective, and cognitive states of the person at the moment of interaction, at fine-grained time scales, from seconds to minutes \cite{d2017advanced}. Activating versus deactivating and positive versus negative emotions are indicators of affective engagement \cite{broughton2010nature}. Although research has shown that positive emotions are more effective than negative emotions in promoting engagement, both positive and negative emotions have the potential to activate engagement \cite{broughton2010nature}. Therefore, engagement cannot be directly derived from high levels of positive or activating emotions. Furthermore, it has been shown that, due to their abrupt changes, six basic emotions, anger, disgust, fear, joy, sadness, and surprise, cannot directly be used as reliable indicators of affective engagement \cite{gupta2016daisee}.

\begin{figure}
    \centering
    \includegraphics[scale=.3]{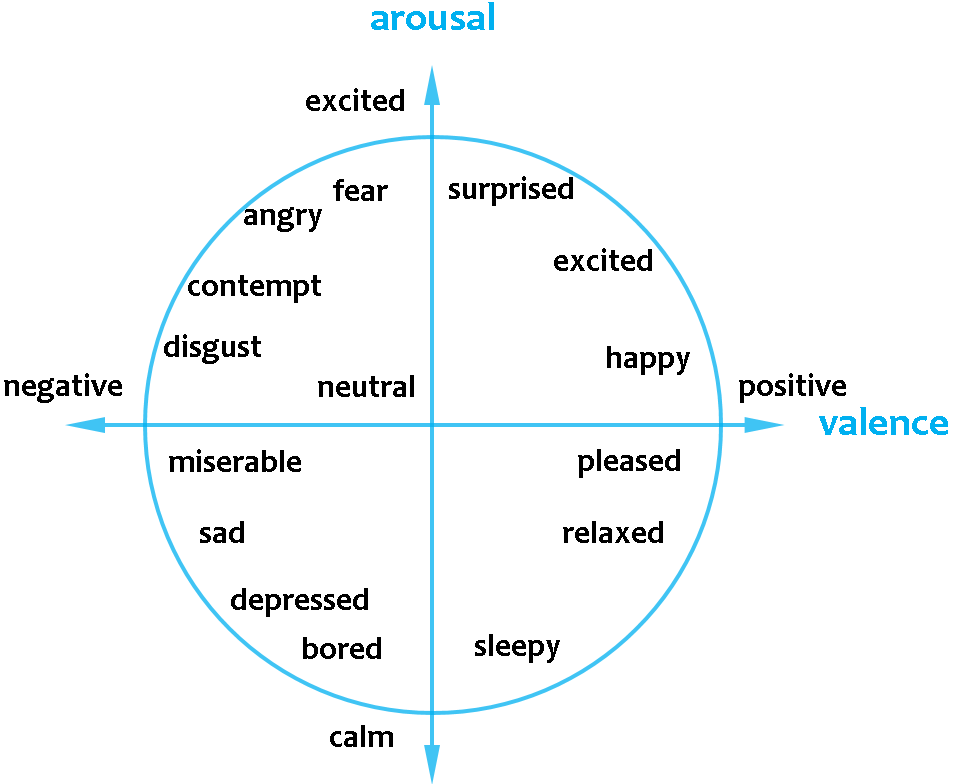}\\
    \caption{The circumplex model of affect \cite{russell1980circumplex}. The positive, and negative values of valence correspond to positive, and negative emotions, respectively, and the positive, and negative values of arousal correspond to activating, and deactivating emotions, respectively.}
    \label{fig:fig1}
\end{figure}

Woolf et al. \cite{woolf2009affect} discretize the affective and behavioral states of persons into eight categories, and defined persons’ desirability for learning, based on positivity or negativity of valence and arousal in the circumplex model of affect \cite{russell1980circumplex} (see Figure \ref{fig:fig1}), and on-task or off-task behavior. In the circumplex model of affect, the positive, and negative values of valence correspond to positive, and negative emotions, and the positive, and negative values of arousal correspond to activating, and deactivating emotions. Persons were marked as being on-task when they are physically involved with the context, watching instructional material, and answering questions, while the off-task behavior represents if the persons are not physically active and attentive to the learning activities. Aslan et al. \cite{aslan2017human} directly defined binary engagement state, engagement/disengagement, as different combinations of positive/negative values of affective states in the circumplex model of affect and on-task/off-task behavior.

In this paper, we assume the only available data modality is a person’s RGB video and propose a video-based person-oriented engagement measurement approach. We train neural network models on affect and behavioral features extracted from videos to automatically infer the engagement level based on the affect and behavioral states of the person in the video.

\subsection{Feature extraction}
\label{sec:feature_extraction}
\begin{figure*}
    \centering
    \includegraphics[scale=.2]{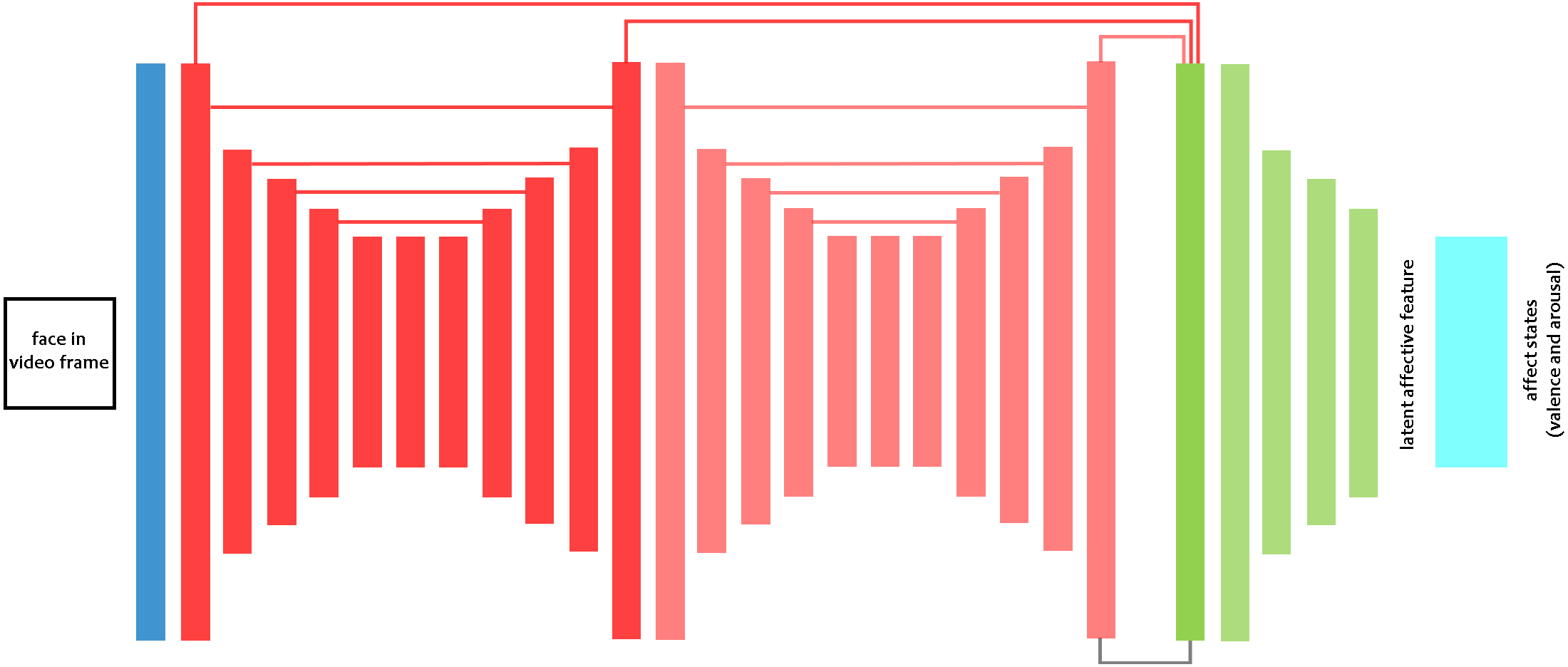}\\
    \caption{The pretrained EmoFAN \cite{toisoul2021estimation} on AffectNet dataset \cite{mollahosseini2017affectnet} is used to extract affect (continuous values of valence and arousal) and latent affective features (a 256-dimensional feature vector). The first blue rectangle and the five green rectangles are 2D convolutional layers. The red and light red rectangles, 2D convolutional layers, constitute two hourglass networks with skip connections. The second hourglass network outputs the facial landmarks that are used in an attention mechanism, the gray line, for the following 2D convolutions. The output of the last green 2D convolutional layer in the pretrained network is considered as the latent affective feature vector. The final cyan fully-connected layer output the affect states.}
    \label{fig:fig2}
\end{figure*}

Toisoul et al. \cite{toisoul2021estimation} proposed a deep neural network architecture, EmoFAN, to analyze facial affect in naturalistic conditions improving the state-of-the-art in emotion recognition, valence and arousal estimation, and facial landmark detection. Their proposed architecture, depicted in Figure \ref{fig:fig2}, is comprised of one 2D convolution followed by two hourglass networks \cite{newell2016stacked} with skip connections trailed by five 2D convolutions and two fully-connected layers. The second hourglass outputs the facial landmarks that are used in an attention mechanism for the following 2D convolutions.

\noindent
\textbf{Affect features:} The continuous values of valence and arousal obtained from the output of the pretrained EmoFAN \cite{toisoul2021estimation} (Figure \ref{fig:fig2}) on the AffectNet dataset \cite{mollahosseini2017affectnet} are used as the affect features.

\noindent
\textbf{Latent affective features:} The pretrained EmoFAN \cite{toisoul2021estimation} on AffectNet \cite{mollahosseini2017affectnet} is used for latent affective feature extraction. The output of the final 2D convolution (Figure \ref{fig:fig2}), a 256-dimensional feature vector, is used as the latent affective features. This feature vector contains latent information about facial landmarks and facial emotions.

\noindent
\textbf{Behavioral features:} The behavioral states of the person in the video, on-task/off-task behavior, are characterized by features extracted from the person’s eye, head, and hand movements.

Ranti et al. \cite{ranti2020blink} demonstrated that blink rate patterns provide a reliable measure of engagement with visual content. They implied that eye blinks are withdrawn at precise moments in time so as to minimize the loss of visual information that occurs during a blink. Probabilistically, the more important the visual information is to the person, the more likely he or she will be to withdraw blinking. We consider blink rate as the first behavioral feature. The intensity of facial Action AU45 indicates how closed the eyes are \cite{baltrusaitis2018openface}. Therefore, a single blink with the successive states of opening-closing-opening appears as a peak in time-series AU45 intensity traces \cite{chen2019faceengage}. The intensity of AU45 in consecutive video frames is considered as the first behavioral feature.

It has been shown that a highly engaged person who is focused on visual content tends to be more static in his/her eye, head, and hand movements and eye gaze direction, and vice versa \cite{sumer2021multimodal, chen2019faceengage}. In addition, in the case of high engagement, the person’s eye gaze direction is towards visual content. Accordingly, inspired by previous research \cite{chen2019faceengage, doherty2018engagement, dewan2019engagement, liao2021deep, whitehill2014faces, niu2018automatic, rivas2021multi, thomas2018predicting, huang2019fine, fedotov2018multimodal, dhall2020emotiw, booth2017toward, kaur2018prediction, wu2020advanced}, eye location, head pose, eye gaze direction, and wrist location in consecutive video frames are considered as the behavioral features.

The features described above are extracted from each video frame and considered as the frame-level features, including:
\begin{itemize}
 \item 2-element affect features (continuous values of valence and arousal),
 \item 256-element latent affective features, and
 \item 12-element behavioral features (eye-closure intensity; \textit{x} and \textit{y} components of eye gaze direction w.r.t. the camera; \textit{x}, \textit{y}, and \textit{z} components of head location w.r.t. the camera; pitch, yaw, and roll as head pose \cite{baltrusaitis2018openface}; \textit{x}, \textit{y}, and \textit{z} components of wrist location w.r.t. the camera).
\end{itemize}

\subsection{Predictive Modeling}
\label{sec:predictive_modeling}
\begin{figure*}
    \centering
    \includegraphics[scale=.14]{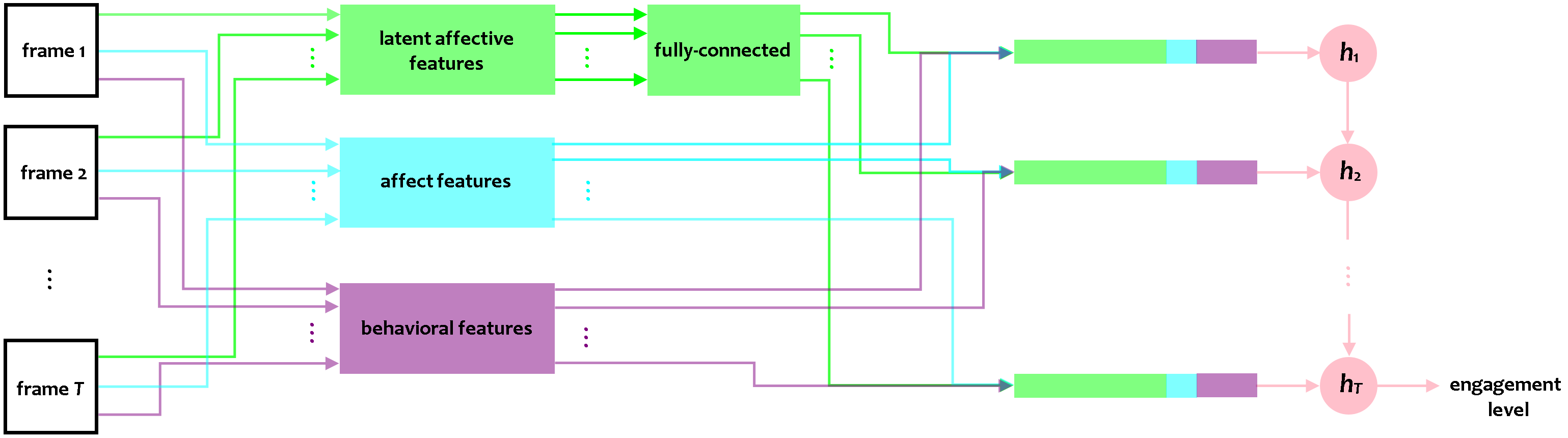}\\
    \caption{The proposed architecture for engagement measurement from video, see Section \ref{sec:predictive_modeling}.}
    \label{fig:fig3}
\end{figure*}

Figure \ref{fig:fig3} shows the structure of the proposed architecture for engagement measurement from video. The latent affective features, followed by a fully-connected neural network, affect features, and behavioral features, are concatenated to construct one feature vector for each video frame. The feature vector extracted from each video frame is considered as one timestamp of a sequential model. The sequential model analyzes the sequences of feature vectors, and the output of the final timestamp of the sequential model, trailed by a fully-connected layer, outputs the engagement level of the person in the video. The fully-connected neural network after latent affective features is jointly trained along with the sequential model and the final fully-connected layer. The fully-connected neural network after the latent affective features is trained to perform joint fusion and to reduce the dimensionality of the latent affective features before being concatenated with the affect and behavioral features. Different sequential models, including Long Short-Term Memory (LSTM) and Temporal convolutional network (TCN) \cite{bai2018empirical}, are investigated in Section \ref{sec:experimental_results}.

\noindent
\textbf{Engagement Measurement through Ordinal Classification:} In ordinal classification, the order of classes is taken into consideration in designing and training machine-learning models \cite{kook2022deep,cardoso2011measuring}. The approach used in this study to incorporate the ordinality of engagement level in developing machine-learning models is as follows.

\textbf{Training–}The original $(K + 1)$-level ordinal labels, $y = 0, 1, …, K$ in the training set are converted into $K$ $y_i$ binary labels as follows, if $y > i: y_i = 0$, else: $y_i = 1$, $i = 0, 1, …, K – 1$. Then, $K$ binary classifiers ($C_i$, $i = 0, 1, …, K – 1$) are trained with the training set and $K$ $y_i$ binary labels.

\textbf{Testing–}For ordinal classification of test samples, each binary classifier $C_i$ gives a probability estimate for each test sample $x_t$, the probability of being in binary class $y_i$, $y_t = y_i$, $i = 0, 1, …, K – 1$. Then, $K$ binary probability estimates are converted into one multi-class probability of being in class $y = 0, 1, …, K$ as follows, 

\begin{eqnarray}\label{eqexpmuts}
p(y_t=k) =
\left\{
	\begin{array}{ll}
		1-p(y_t>0),  & \mbox{if }  k=0 \\
		p(y_t>k-1)-p(y_t>k), & \mbox{if } 0<k<K-1 \\
		p(y_t>K-2), & \mbox{if } k=K-1
	\end{array}
\right..
\end{eqnarray}

\section{Experimental Results}
\label{sec:experimental_results}
In this section, the performance of the proposed engagement measurement method is evaluated compared to the previous feature-based and end-to-end methods. The classification and regression results on two publicly available video-based engagement datasets are reported. The ablation study of different feature sets is studied, and the effectiveness of affect states in engagement measurement is investigated. Finally, the impact of incorporating the ordinality of engagement in model training is investigated through comparing the results of ordinal and non-ordinal models.

\subsection{Datasets}
The performance of the proposed method is evaluated on the only two publicly available video-only engagement datasets, DAiSEE \cite{gupta2016daisee} and EmotiW-EW \cite{kaur2018prediction}. 

\noindent
\textbf{DAiSEE:} The DAiSEE dataset \cite{gupta2016daisee} contains 9,068 videos captured from 112 persons in online courses. The videos were annotated by four states of persons while watching online courses, boredom, confusion, frustration, and engagement. Each state is in one of the four levels (ordinal classes), level 0 (very low), 1 (low), 2 (high), and 3 (very high). In this paper, the focus is only on the engagement level classification. The length, frame rate, and resolution of the videos are 10 seconds, 30 frames per second (fps), and 640 × 480 pixels. Table \ref{tab:tab2} shows the distribution of samples in different classes and the number of persons in the train, validation, and test sets. These sets are used in our experiments to fairly compare the proposed method with the previous methods. The results are reported on 1784 test videos. As can be seen in Table \ref{tab:tab2}, the dataset is highly imbalanced.

\noindent
\textbf{EmotiW-EW:} The EmotiW-EW dataset has been released in the student engagement measurement sub-challenge of the Emotion Recognition in the Wild Challenge \cite{kaur2018prediction}. It contains videos of 78 persons in online classroom settings. The total number of videos is 262, including 148 training, 48 validation, and 67 test videos. The videos are at a resolution of 640 × 480 pixels and 30 fps. The lengths of the videos are around 5 minutes. Each video has been annotated as having one of four engagement levels, 0.00, 0.33, 0.66, or 1.00, where 0.00, and 1.00 indicate the person is completely disengaged, and highly engaged, respectively. In this sub-challenge, the engagement measurement has been defined as a regression problem, and only training and validation sets are publicly available. We use the training, and validation sets for training, and validating the proposed method, respectively. The distribution of samples in this dataset is also imbalanced, Table \ref{tab:tab3}.

\begin{table}[ht]
\caption{The distribution of samples in different engagement level classes, and the number of persons, in train, validation, and test sets in the DAiSEE dataset \cite{gupta2016daisee}.}
\label{tab:tab2}
\centering
\begin{tabular}{p{.2\linewidth}p{.15\linewidth}p{.15\linewidth}p{.15\linewidth}}
\hline
level & train & validation & test\\
\hline
0 & 34 & 23 & 4\\
\hline
1 & 213 & 143 & 84\\
\hline
2 & 2617 & 813 & 882\\
\hline
3 & 2494 & 450 & 814\\
\hline
total & 2358 & 1429 & 1784\\
\hline
\# of persons & 70 & 22 & 20\\
\hline
\end{tabular}
\end{table}

\begin{table}[ht]
\caption{The distribution of samples in different engagement level values in the train and validation sets of the EmotiW-EW dataset \cite{kaur2018prediction}.}
\label{tab:tab3}
\centering
\begin{tabular}{p{.15\linewidth}p{.15\linewidth}p{.15\linewidth}}
\hline
level & train & validation\\
\hline
0.00 & 6 & 4\\
\hline
0.33 & 35 & 10\\
\hline
0.66 & 79 & 19\\
\hline
1.00 & 28 & 15\\
\hline
total & 148 & 48\\
\hline
\end{tabular}
\end{table}

\subsection{Experimental Setting}
\label{sec:experimental_setting}
The behavioral eye and head features, and hand features (described in Section \ref{sec:feature_extraction}) are extracted by the OpenFace \cite{baltrusaitis2018openface}, and MediaPipe \cite{lugaresi2019mediapipe}, respectively. The OpenFace also outputs the extracted face regions from video frames. The extracted face regions of size 256 × 256 are fed to the pretrained EmoFAN \cite{toisoul2021estimation} on AffectNet \cite{mollahosseini2017affectnet} for extracting affect features and latent affective features \ref{sec:feature_extraction}.

The fully-connected neural network after the latent affective features in Figure \ref{fig:fig3} contains two fully-connected layers with 256 × 128, and 128 × 32 neurons. The output of this fully-connected neural network, a 32-element reduced dimension latent affective feature vector, is concatenated with the affect and behavioral features to generate a 46 (32 + 2 + 12)-element feature vector for each video frame and correspondingly for each timestamp of the sequential model. We have implemented two sequential models for analyzing the extracted features and outputting the engagement level, including LSTM and TCN. The LSTM has two unidirectional layers with 46 × 128 and 128 × 64 layers, followed by a fully-connected layer at its final timestamp with 64 × 4 neurons for 4-class classification. The parameters of the TCN, giving the best results, are as follows, 8, 128, 16, and 0.25 for the number of levels, number of hidden units, kernel size, and dropout \cite{bai2018empirical}. At the final timestamp of the TCN, a fully-connected layer with 4 output neurons is used for 4-class classification (in the DAiSEE dataset).

The EmotiW-EW dataset contains videos of around 5-minute length. The videos are divided into 10-second clips with 50\% overlap \cite{thomas2018predicting}, and clip-level features are extracted from each clip as follows. First, affect states and behavioral features, described in Section \ref{sec:feature_extraction}, are extracted from consecutive video frames in each clip. Then, for each clip, a 49-element clip-level feature vector is created as follows.
\begin{itemize}
  \item 4 features: the mean and standard deviation of valence and arousal values over consecutive clip frames,
  \item 1 feature: the blink rate, derived by counting the number of peaks above a certain threshold divided by the number of frames in the AU45 intensity time series extracted from the input clip,
  \item 8 features:  the mean and standard deviation of the velocity and acceleration of x and y components of eye gaze direction,
  \item 12 features:  the mean and standard deviation of the velocity and acceleration of x, y,  and z components of head location,
  \item 12 features:  the mean and standard deviation of the velocity and acceleration of head’s pitch, yaw, and roll, and
  \item 12 features:  the mean and standard deviation of the velocity and acceleration of x, y,  and z components of wrist location.
 \end{itemize}

The sequence of clip-level features is analyzed by a two-layer unidirectional LSTM with 49 × 128 and 128 × 64 layers, followed by a fully-connected layer at its final timestamp with 64 × 1 neurons for regression in the EmotiW-EW dataset. In addition, a TCN with 8, 128, 16, and 0.25 for the number of levels, number of hidden units, kernel size, and dropout \cite{bai2018empirical} has also been implemented. At the final timestamp of the TCN, a fully-connected layer with one output neuron is used for regression in the EmotiW-EW dataset.

The evaluation metrics are classification accuracy and Mean Squared Error (MSE) for the regression task. The experiments were implemented in PyTorch \cite{paszke2019pytorch} and Scikit-learn \cite{pedregosa2011scikit} on a  server with  64  GB  of  RAM  and  NVIDIA  TeslaP100  PCIe  12  GB  GPU.

\subsection{Results}
\label{sec:results}
Table \ref{tab:tab4} shows the engagement level classification accuracy of different settings of the proposed method on the test set of the DAiSEE dataset. As ablation study, in the upper section of Table \ref{tab:tab4}, the results of different classification models are reported using different feature sets. In both LSTM and TCN, adding behavioral and affect features to the latent affective features improves the accuracy, showing the effectiveness of affect states in engagement measurement. The TCN can achieve higher accuracy than the LSTM because of its superiority in modeling sequences of longer lengths and its ability to retain memory of history. The lower section of Table \ref{tab:tab4} shows the results of ordinal LSTM and ordinal TCN using all the features. As described in Section \ref{sec:predictive_modeling}, three binary classifiers are trained and used for inference for the 4-level ordinal engagement level classification problem in the DAiSEE dataset. Comparing the non-ordinal and ordinal versions of the LSTM and TCN (using all the features) in the upper and lower sections of Table \ref{tab:tab4}, respectively, shows the superiority of the ordinal classifiers. The highest classification accuracy (67.4\%) was achieved by the ordinal TCN with all the latent affective, behavioral, and affect features. Table \ref{tab:tab5} shows the engagement level classification accuracy of the previous methods on the test set of the DAiSEE dataset. According to Table \ref{tab:tab4} and \ref{tab:tab5}, the proposed method (ordinal TCN with all the features) outperforms all the previous methods.

\begin{table}[ht]
\caption{Engagement level classification accuracy of different settings of the proposed method on the test set of the DAiSEE dataset. LSTM: long short-term memory, TCN: temporal convolutional network.}
\label{tab:tab4}
\centering
\begin{tabular}{p{.5\linewidth}p{.2\linewidth}p{.15\linewidth}}
\hline
feature set & model & accuracy\\
\hline
latent affective & LSTM & 60.8\\
\hline
latent affective + behavioral & LSTM & 62.3\\
\hline
latent affective + behavioral + affect & LSTM & 63.9\\
\hline
latent affective & TCN & 60.2\\
\hline
latent affective + behavioral & TCN & 62.5\\
\hline
latent affective + behavioral + affect & TCN & 64.7\\
\hline
\hline
latent affective + behavioral + affect & Ordinal LSTM & 65.6\\
\hline
\textbf{latent affective + behavioral + affect} & \textbf{Ordinal TCN} & \textbf{67.4}\\
\hline
\end{tabular}
\end{table}

\begin{table}[ht]
\caption{Engagement level classification accuracy of the previous methods on the DAiSEE dataset (see Section \ref{sec:literature_review}).}
\label{tab:tab5}
\centering
\begin{tabular}{p{.5\linewidth}p{.15\linewidth}}
\hline
method & accuracy\\
\hline
C3D \cite{gupta2016daisee} & 48.1\\
\hline
I3D \cite{zhang2019novel} & 52.4\\
\hline
C3D + LSTM \cite{abedi2021improving} & 56.6\\
\hline
C3D with transfer learning \cite{gupta2016daisee} & 57.8\\
\hline
LRCN \cite{gupta2016daisee} & 57.9\\
\hline
DFSTN \cite{liao2021deep} & 58.8\\
\hline
C3D + TCN \cite{abedi2021improving} & 59.9\\
\hline
DERN \cite{huang2019fine} & 60.0\\
\hline
ResNet + LSTM \cite{abedi2021improving} & 61.5\\
\hline
Neural Turing Machine \cite{ma2021automatic} & 61.3\\
\hline
3D DenseNet \cite{mehta2022three} & 63.6\\
\hline
ResNet + TCN \cite{abedi2021improving} & 63.9\\
\hline
ShuffleNet v2 \cite{hu2022optimized} & 63.9\\
\hline
\end{tabular}
\end{table}

According to Abedi and Khan \cite{abedi2021improving} and Liao et al. \cite{liao2021deep}, due to highly imbalanced data distribution (see Table \ref{tab:tab2}), none of the previous approaches on the DAiSEE dataset (outlined in Table \ref{tab:tab5}) are able to correctly classify samples in the minority classes (classes 0 and 1). Table \ref{tab:tab6} shows the confusion matrices of one of the previous end-to-end approaches (a) and three different settings of the proposed feature-based approach (b)-(d) with superior performance on the test set of the DAiSEE dataset. Comparing Tables \ref{tab:tab6} (c) and (d) with (b) shows the effectiveness of adding affect state features and using ordinal classification settings in the correct classification of samples in the minority classes 0 and 1. Specifically, the ordinal setting in (d) compared to the regular setting in (c) significantly increases the number of samples in the main diagonal of the confusion matrix, i.e., the number of correctly classified samples.

\begin{table}[ht]
\caption{The confusion matrices of (a) an end-to-end approach, ResNet + TCN \cite{abedi2021improving} and the proposed feature-based approaches on the test of the DAiSEE dataset, (b) (latent affective + behavioral) features + TCN, (c) (latent affective + behavioral + affect) features + TCN, and (d) (latent affective + behavioral + affect) features + Ordinal TCN.}
\label{tab:tab6}
\centering
\includegraphics[scale=.3]{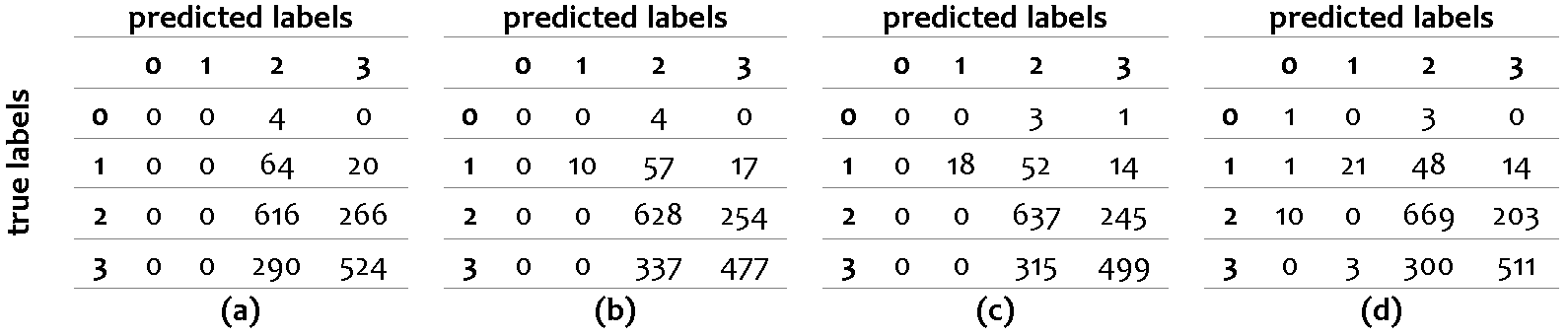}
\end{table}

\begin{figure}
    \centering
    \includegraphics[scale=.22]{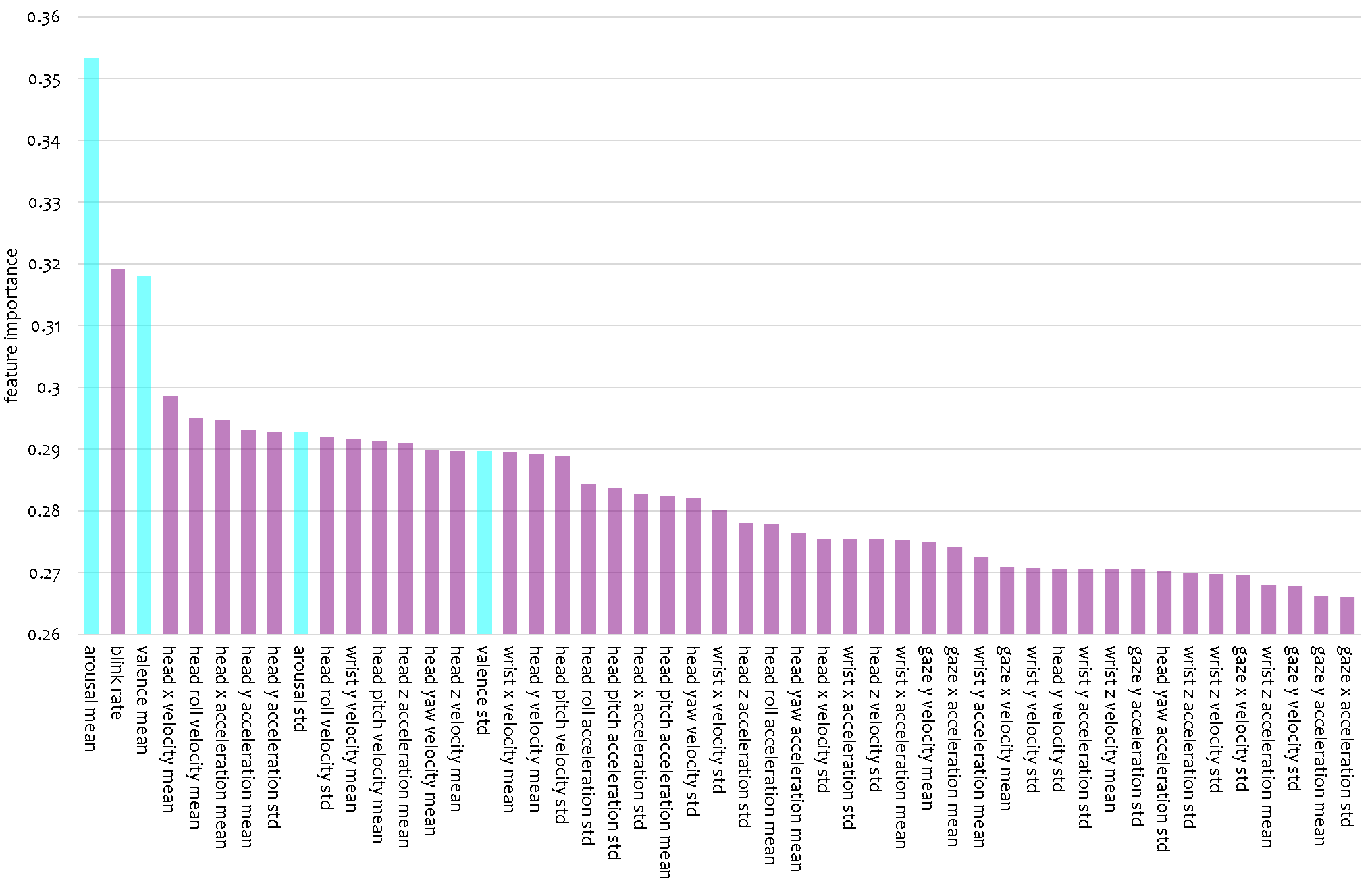}
    \caption{Features importance ranking of clip-level features using Random Forest (see Section \ref{sec:results}).}
    \label{fig:fig4}
\end{figure}

Figure \ref{fig:fig4} shows the importance of affect and behavioral features using a random forest classifier by checking their out-of-bag error \cite{chen2019faceengage}. One set of clip-level features (described in Section \ref{sec:experimental_setting}) is extracted from each video sample of the DAiSEE dataset and used as the input to the random forest for feature importance ranking. The feature importance ranking in Figure \ref{fig:fig4} is consistent with evidence from psychology and physiology. (i) The first, and third indicative features are the mean values of arousal, and valence, respectively. It aligns with the findings in \cite{woolf2009affect, aslan2017human} indicating that affect states are highly correlated with the desirability of content, concentration, satisfaction, excitement, and affective engagement. (ii) The second important feature is the blink rate. It aligns with the research by Ranti et al. \cite{ranti2020blink} showing that blink rate patterns provide a reliable measure of individual behavioral and cognitive engagement with visual content. (iii) The least important features are extracted from eye gaze direction. Research conducted by Sinatra et al. \cite{sinatra2015challenges} shows that to have a high impact on engagement measurement, eye gaze direction features should be complemented by contextual features that encode general information about the changes in visual content.

Table \ref{tab:tab7} shows the regression MSE of applying different methods to the validation set in the EmotiW-EW dataset. All the outlined methods in Table \ref{tab:tab7} are feature-based, described in Section \ref{sec:literature_review}. In the proposed method, the clip-level features (see Section \ref{sec:literature_review}) are extracted from video samples and used as the input to the consecutive timestamps of the sequential models for regression. Adding the affect features to the behavioral features reduces MSE (second and fourth rows of Table \ref{tab:tab7}), showing the effectiveness of the affect states in engagement level regression. The best setting of the proposed method, including the behavioral and affect features and TCN, outperforms all the previous methods except \cite{copur2022engagement}.

\begin{table}[ht]
\caption{Engagement level regression MSE of applying different methods on the validation set of the EmotiW-EW dataset.}
\label{tab:tab7}
\centering
\begin{tabular}{p{.8\linewidth}p{.1\linewidth}}
\hline
method & MSE\\
\hline
clip-level behavioral features + LSTM (proposed) & 0.06230\\
\hline
clip-level (behavioral + affect) features + LSTM (proposed) & 0.0597\\
\hline
clip-level behavioral features + TCN (proposed) & 0.0618\\
\hline
\textbf{clip-level (behavioral + affect) features + TCN (proposed)} & \textbf{0.0508}\\
\hline
\hline
eye and head-pose features + LSTM \cite{dhall2020emotiw} & 0.1000\\
\hline
DFSTN \cite{liao2021deep} & 0.0736\\
\hline
eye, head-pose, and AU features + TCN \cite{thomas2018predicting} & 0.0655\\
\hline
eye, head-pose, and AU features + GRU \cite{niu2018automatic} & 0.0671\\
\hline
\textbf{eye, head-pose, and AU features + LSTM \cite{niu2018automatic}} & \textbf{0.0427}\\
\hline
\end{tabular}
\end{table}

\subsection{An Illustrative Example}
\begin{figure*}
    \centering
    \includegraphics[scale=.25]{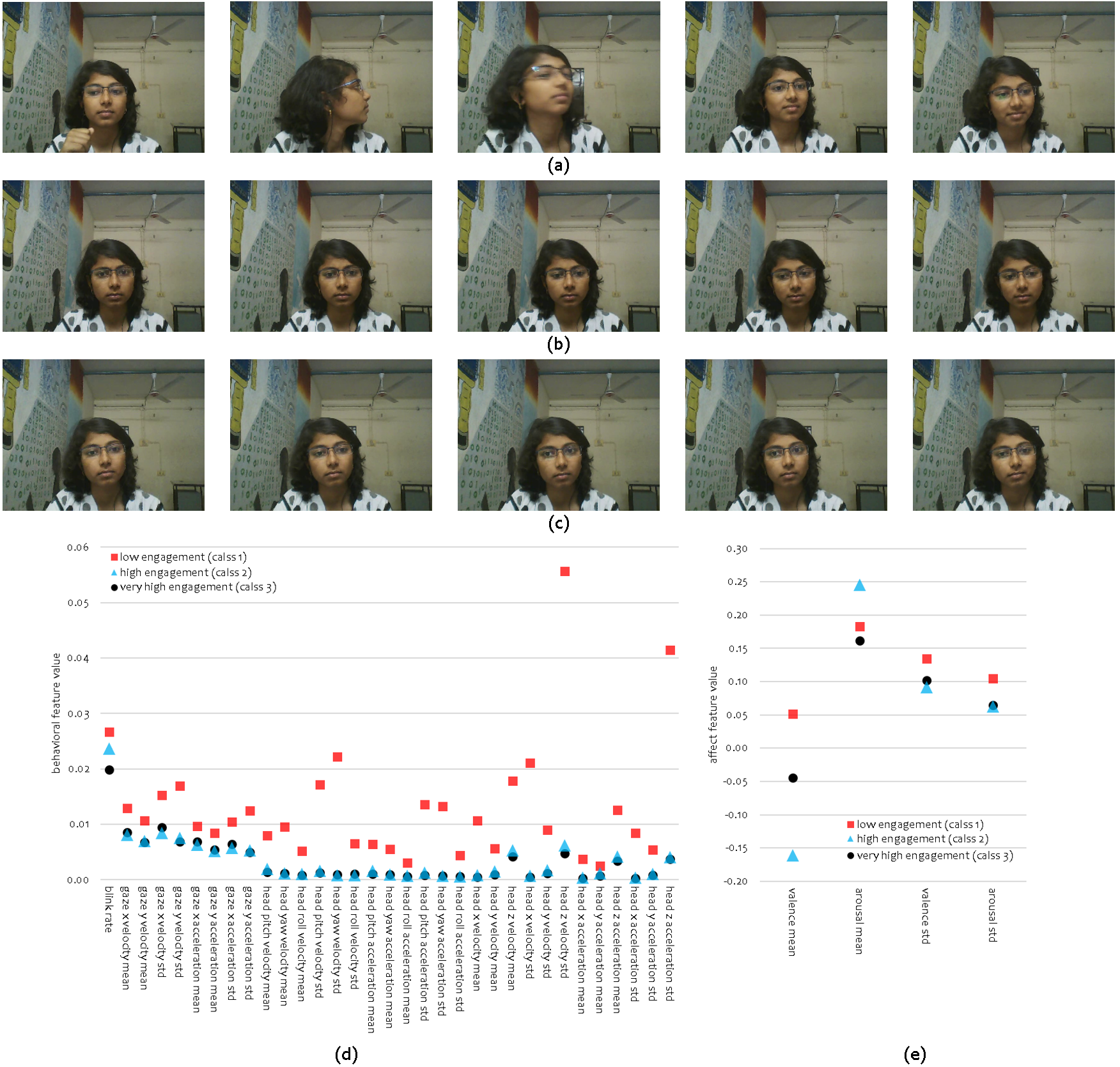}
    \caption{An illustrative example for showing the importance of affect states in engagement measurement. 5 (of 300) frames of three different videos of one person in the DAiSEE dataset in classes (a) low, (b) high, and (c) very high levels of engagement, the clip-level (d) behavioral features and (e) affect features of the three videos. In (a), the person does not look at the camera for a moment and behavioral features in (d) are totally different for this video compared to the videos in (b) and (c). However, the behavioral features for (b) and (c) are almost identical. According to (e), the affect features are different for (b) and (c) and are effective in differentiation between the videos in (b) and (c).}
    \label{fig:fig5}
\end{figure*}

Figure \ref{fig:fig5} shows 5 (of 300) frames of three different videos of one person in the DAiSEE dataset. The videos in (a), (b), and (c) are in low (1), high (2), and very high (3) levels of engagement. Figure \ref{fig:fig5} (d), and (e) depict the clip-level eye and head behavioral features, and affect features of these three videos, respectively. As can be observed in Figure \ref{fig:fig5} (d), the behavioral features of the two videos in classes 2 and 3 are different from the video in class 1. Therefore, the behavioral features can differentiate between classes 2 and 3, and class 1. While the values of the behavioral features for the two videos in classes 2 and 3 are almost identical, according to Figure \ref{fig:fig5} (e), the mean values of valence and arousal are different for these two videos. According to this example, the confusion matrices, and the accuracy and MSE results in Tables 4-7, affect features are necessary to differentiate between different levels of engagement.

\subsection{Discussion}
\label{sec:discussion}
The proposed method achieved superior performance compared to the previous feature-based and end-to-end methods on the DAiSEE dataset, particularly in terms of classifying low levels of engagement. However, the overall classification accuracy on this dataset needs to be improved. This can be due to the following two reasons. The first reason is the imbalanced distribution of data, with a relatively small number of samples in low levels of engagement when compared to high levels of engagement, Table \ref{tab:tab2}. During training, with a conventional shuffled dataset, there will be many training batches that do not include samples with low levels of engagement. We tried to mitigate this problem with a sampling strategy in which the samples of all classes were included in each batch. By doing so, the model will be trained on samples of all classes in each training iteration. The second reason is the annotation problems in the DAiSEE dataset. According to our study on the videos in the DAiSEE dataset, there are annotation mistakes in this dataset. These mistakes are more obvious when comparing videos and annotations of one person in different classes, as it is discussed with examples by Liao et el. \cite{liao2021deep} and Mehta et al. \cite{mehta2022three}. Obviously, the default reason for not achieving greater classification accuracy is the difficulty of this classification task, the small differences between videos at different levels of engagement, and the lack of any other type of data other than video.

\section{Conclusion and Future Work}
\label{sec:conclusion}
In this paper, we proposed a novel method for objective engagement measurement from videos of a person watching an online course. The experiments were performed on the only two publicly available engagement measurement datasets (DAiSEE \cite{gupta2016daisee} and EmotiW-EW \cite{kaur2018prediction}), containing only video data. As any complementary information about the context (e.g., the degree level of the students, the lessons being taught, their progress in learning, and the speech of students) was not available, the proposed method is a person-oriented engagement measurement using only video data of persons at the moment of interaction. Again, due to the lack of context (e.g., students’ and tutors’ speech), their cognitive engagement could not be measured. Therefore, we were only able to measure affective and behavioral engagements based on visual indicators (see Section \ref{sec:what_is_engagement}). We used affect sates, continuous values of valence and arousal extracted from consecutive video frames, along with a new latent affective feature vector and behavioral features for engagement measurement. For the purpose of predicting engagement as an ordinal variable, we developed ordinal classification models. The proposed affect-driven ordinal approach achieved superior performance in terms of classification accuracy and correctly classifying disengaged videos in the DAiSEE dataset, as well as very low regression MSE on the validation set of the EmotiW-EW dataset.

In future work, we plan to collect an audio-video dataset from patients in virtual rehabilitation sessions. We will extend the proposed video-based engagement measurement method to work in tandem with audio. We will incorporate audio data, and also context information (progress of patients in the rehabilitation program and rehabilitation program completion barriers) to measure engagement at a deeper level, e.g., cognitive and context-oriented engagement (see Section \ref{sec:what_is_engagement}). In addition, we will explore approaches to measure affect states from audio and use them for engagement measurement. To avoid any mistakes in engagement annotation, as it was discussed in Section \ref{sec:discussion}, we will use psychology-backed measures of engagement \cite{khan2022inconsistencies}. Another line of our future research will focus on detecting disengagement as an anomaly using autoencoders and contrastive learning techniques \cite{khosla2020supervised}.
\\\\
\textbf{Data availability}\\
The datasets analyzed during the current study are publicly available in the following repositories:\\
https://people.iith.ac.in/vineethnb/resources/daisee/index.html\\
https://sites.google.com/view/emotiw2020/
\\\\
\textbf{Conflict of Interest}\\
The authors declare that they have no conflict of interest.

\bibliography{sn-bibliography}
\end{document}